\pdfoutput=1

\documentclass[11pt]{article}

\usepackage{acl}

\usepackage{times}
\usepackage{latexsym}
\usepackage{booktabs}
\usepackage{mathtools}
\usepackage{amsmath}
\usepackage{bbm}
\usepackage{subcaption}
\usepackage{amsthm}
\usepackage{float}
\usepackage[T1]{fontenc}

\usepackage[utf8]{inputenc}

\usepackage{microtype}

\usepackage{inconsolata}

\usepackage{graphicx}

\usepackage{hyperref}

\title{Metric assessment protocol\\in the context of answer fluctuation on MCQ tasks}

\author{
  \textbf{Ekaterina Goliakova \textsuperscript{1,2}},
  \textbf{Xavier Renard\textsuperscript{1,2}},
  \textbf{Marie-Jeanne Lesot\textsuperscript{1}},
  \textbf{Thibault Laugel\textsuperscript{1,2}},
\\  \textbf{Christophe Marsala\textsuperscript{1}},
  \textbf{Marcin Detyniecki\textsuperscript{1,2,3}}
 \\
 \\
 \textsuperscript{1}Sorbonne University, CNRS, LIP6, Paris, France \\
  \textsuperscript{2}AXA, Paris, France \\
  \textsuperscript{3}Polish Academy of Science, IBS PAN, Warsaw, Poland 
\\
  \small{
    \textbf{Correspondence:} \href{mailto:email@domain}{ekaterina.goliakova@lip6.fr}
  }
}
\begin{document}
\maketitle
\begin{abstract}

Using multiple-choice questions (MCQs) has become a standard for assessing LLM capabilities efficiently. 
A variety of metrics can be employed for this task. However, previous research has not conducted a thorough assessment of them. At the same time, MCQ evaluation suffers from \textit{answer fluctuation}: models produce different results given slight changes in prompts. We suggest a metric assessment protocol in which evaluation methodologies are analyzed through their connection with fluctuation rates, as well as original performance. Our results show that there is a strong link between existing metrics and the answer changing, even when computed without any additional prompt variants. A novel metric, \textit{worst accuracy}, demonstrates the highest association on the protocol.

\end{abstract}

\section{Introduction}

Testing on question answering tasks has become standard in the LLM evaluation field \cite{rogers2023qa}. 
However, assessing models' generations in these conditions is a complex task, due to inapplicability of "traditional"  metrics, such as BLEU \cite{papineni-etal-2002-bleu}, ROUGE \cite{lin-2004-rouge}, or BERTScore \cite{Zhang2019BERTScoreET}, because of high variation between possible correct answers \cite{he2022blind, sulem2018bleu}. While human evaluation can be used instead, it can be costly \cite{elangovan2024considers} and subjective \cite{elangovan2024beyond, abeysinghe2024challenges}. 
Thus, multiple-choice questions (MCQ) benchmarks have prevailed in LLM evaluation, as a tool that maps all possible responses to a small set of options, with examples such as ARC \cite{clark2018think}, GPQA \cite{rein2024gpqa}, and BigBench-Hard \cite{suzgun2022challenging}.

Using MCQ tasks allows for the exact matching of answers selected by models and correct ones and for the computation of standard metrics, such as accuracy \cite{gemmateam2024gemma2improvingopen, achiam2023gpt, wang2024mmlu}. While reporting accuracy is typical, the metrics available for MCQ tasks include other possibilities. 
For instance, continuous metrics such as \textit{probability mass} of correct answer can improve signal-to-noise ratio in evaluations \cite{madaan2024quantifying} or better track actual performance of models of different sizes during training \cite{schaeffer2023emergent, du2024understanding}. Additionally, new metrics were proposed specifically in the context of MCQ evaluation (e.g. \citealp{pezeshkpour2023large, zheng2023mcq}). However, previous work has not provided a thorough comparative analysis of these metrics.

In addition, prior research \cite{pezeshkpour2023large, gupta2024changing, li2024anchored, zheng2023mcq, tjuatja2024llms} indicates that LLMs are sensitive to changes in MCQ options order: it is possible to elicit a different response from a model simply by rearranging the proposed answers. The phenomenon of LLMs producing different answers given semantically insignificant prompt changes can be called \textit{answer fluctuation} (\citealp{wei2024unveiling}) or answer floating (\citealp{wang2024look}).

A deep understanding of answer fluctuation is crucial since LLMs' reliability remains a concern, especially in sensitive domains \cite{khatun-brown-2023-reliability, amiri2024enhancing, naik2024probabilistic}. Nevertheless,
discovering all cases of fluctuation leads to significantly higher computation costs, due to the necessity of testing multiple prompts.  

We propose to use this factor in order to compare metrics available for the evaluation of MCQ tasks. In particular, we perform the costly calculation of models' responses fluctuation on all possible permutations and then compare those results with metrics computed on smaller subsets of permutations, assessing if any of the metrics could be used as a cost-efficient proxy for the \textit{full fluctuation rates} (computed on all permutations), without losing the information about the original performance.  Our contributions can be summarized as follows:
\begin{enumerate}
    \item Compilation and formalization of existing metrics used for estimating LLMs' performance on MCQ benchmarks (Section \ref{metric_survey}).
    \item Proposition of a novel metric for MCQ evaluation (Section~\ref{metric_proposition}).
    \item Introduction of a metric assessment protocol in which we analyze how well a given metric correlates with full fluctuation rates, as well as the original accuracy of the model (Section~\ref{protocol}).
    \item Application of the protocol to the results of 10 models on 17 tasks (Section \ref{results}).
\end{enumerate}

We find that most metrics strongly correlate with the full fluctuation rates, even when calculated only on the original version of the benchmark. However, the correlation becomes stronger when adding results from multiple permutations, achieving the coefficient of determination $R^2 > 0.9$ for \textit{partial fluctuation rates} (computed on subsets of permutations) and the novel metric, \textit{worst accuracy}.

\section{Context \& Related Work} \label{related_work}
MCQs have been widespread in the education field \cite{brady2005assessment, moss2001multiple}. They are characterized by presenting several answer \emph{options} within a question body, typically accompanied by \emph{labels} (e.g. \texttt{A/B/C/D}), where a \emph{correct answer} can be one, several, or no labels. In the context of LLMs evaluation, however, MCQ benchmarks come with a single correct label, see an example in Figure \ref{faketable:mcq}. The unique correct answer allows for comparing models' responses to it and obtaining accuracy.

As for the extraction of a model's responses, one can compare probabilities of the next token given a question prompt and choose the most probable one as the model's selected label. Another method prominent in the field, though not covered in this paper, is to allow models to generate an answer of arbitrary length and later classify it as one of the labels \cite{wang2024my}.
\begin{figure}[t]
\begin{center}
\begin{tabular}{c|c}
\toprule
\multicolumn{2}{c}{Which of these will form new soil the fastest?}\\
\toprule
\textbf{Labels} & \textbf{Options} \\
\toprule
\textbf{A} & A log rotting in a forest. \\
\midrule
\textbf{B} & Water running in a stream. \\
\midrule
\textbf{C} & A rock sitting in a garden. \\
\midrule
\textbf{D} & Waves breaking on a beach. \\
\bottomrule
\multicolumn{2}{c}{Correct label: \textbf{A}}\\
\bottomrule
\end{tabular}
\end{center}
\caption{An MCQ example from ARC-C \cite{clark2018think}.} \label{faketable:mcq}
\end{figure}

Previous research demonstrates that one can cause answer fluctuation by permuting questions, their options and/or labels.

\paragraph{Answer fluctuation}
\citealp{mizrahi2024state} show that even minimal prompt paraphrases, e.g., replacing "have" with "include" in the question, impact models' performance. 
\citealp{liang2022holistic} indicate that a different choice of few-shot examples can lead to vast differences in obtained F1 scores. \citealp{mina2025cognitive}, as well, highlight the effect of few-shot examples, where recency bias (preference towards selecting the last option) is found in the few-shot scenario but not the zero-shot scenario.

\citealp{pezeshkpour2023large} study the effect of option order permutation. Their work shows that the difference between the best and worst possible performance of a model achievable via option reordering can be as high as 70 percentage points for InstructGPT and 50 percentage points for GPT-4, highlighting the fact that the introduction of few-shot examples does not lead to higher robustness. 

\citealp{zheng2023mcq} demonstrate that moving all correct answers to one of \texttt{A/B/C/D} can cause a performance increase in some models and a decrease in others, serving as an example of \textit{selection bias} \cite{li2024anchored, pezeshkpour2023large, wang2023large}. 
Additionally, using different option typography (e.g., \texttt{(A)} instead of \texttt{A.} or replacing common option labels \texttt{A/B/C/D} with rarer ones, e.g. \texttt{\$/\&/\#/@}) leads to lower results \cite{zheng2023mcq, alzahrani2024benchmarks}. Furthermore, a similar drop in performance is achieved \cite{wei2024unveiling} if one keeps the order of options but reverses the order of labels (e.g., \texttt{D/C/B/A}).  

\citealp{tjuatja2024llms} compare LLMs' biases on MCQ with those of people and find no apparent replication of human behavior, while indicating that all tested models show sensitivity to factors not significant for human respondents, such as typos.


Finally, changing the question from MCQ to another format, such as Cloze \cite{madaan2024quantifying}, open-ended generation \cite{rottger2024political}, or True/False questions \cite{wang2024leastincorrect} can drastically change models' responses.

\paragraph{LLM evaluation in the fluctuation context}
Given the answer instability, \citealp{wei2024unveiling} propose the \textit{fluctuation rates} metric that compares answers on the original and inverse option orders. It considers that a model's response fluctuates if these answers are different. However, this calculation is not adapted for working with multiple permutations.

To ensure more stable model performance, \citealp{zheng2023mcq} introduce \textit{PriDe} (\citealp{li2024calibraeval, wei2024unveiling, reif2024beyond} present other calibration techniques): an approach to adjust models' probabilities of answer tokens (e.g. \texttt{A/B/C/D}) by computing their priors, independent from questions, and then using them to debias models' responses. This methodology has only been evaluated in terms of improving the original performance of models,  not considering the evaluation of answer robustness.

\textit{Sensitivity gap} \cite{pezeshkpour2023large} is one of the proposed metrics that incorporates the information about both model performance and answer fluctuation. It is computed as the difference between the maximum and minimum accuracies that can be obtained by changing the order of options. However, the paper does not provide the exact formula for this calculation. Similarly, \citealp{gupta2024changing} introduce an unnamed metric to assess, which we take the liberty to name  \textit{strong accuracy}. It compares pair-wise responses from the original option order and a permutation and calculates an average rate of keeping correct answers through permutation pairs. Their approach involves picking random permutations, although the stability of the metric is not addressed.

To the best of our knowledge, the above-mentioned metrics have not been substantially compared to one another, as well as to robustness. 
The connection of reliability and other metrics has remained underexplored, being demonstrated only for accuracy \cite{pezeshkpour2023large, liang2022holistic, wei2024unveiling}.





\section{Metrics Survey} \label{metric_survey}
Given the variety of metrics available for MCQ evaluation, it is essential to provide a coherent formalization for each of them. This section presents our notation and permutation types used for computation. Furthermore, we provide formulas for existing metrics. Finally, we introduce a novel metric, that we call \textit{worst accuracy}.
\subsection{Notation} 
We assume that all benchmarks come with their own set of labels $L$ (such as \texttt{A/B/C/D}), as well as a set of questions. We define each
 metric for a question $q$ and, within our experiments, we average all calculations among questions. However, one can potentially adopt different aggregation strategies. 
 
 Each question has an associated set of textual options $O = \{o_1 \dots o_{|L|}\}$, e.g. $\{cat, dog \dots  \}$, as well as a correct answer $a$ (e.g. $dog$).
 We define a permutation set $\mathcal{R}(O)$ as a set of reordering of set $O$, e.g. $\mathcal{R}(O) = \{\{o_1, o_2, o_3, o_4\}, \{o_4, o_3, o_2, o_1\}\}$. Given few-shot examples, question $q$ and permuted options $r_{j} \in \mathcal{R}(O)$, we obtain model answer $m_{j}$. 

 Please note that the labels are not permuted. Therefore, a label of the correct answer might differ among permutations. To keep track of it, we introduce the notation $l_{a_j}$ which stands for the label of the correct answer $a$ on a permutation $ r_j \in \mathcal{R}(O)$. Few-shot examples and the question itself remain constant throughout the permutations, and for this reason, they are not presented in subsequent formalization.


\subsection{Permutation types}
When all possible orders of options are present, we call such a permutation set $\mathcal{R}_{full}$. Since $|\mathcal{R}_{full}|=|L|!$, its calculation is extremely costly. To make computations more efficient, we employ subsets of permutations. 

If the permutation set contains only the original options order, we call refer to it as $\mathcal{R}_{original}$.
Previous research \cite{wei2024unveiling}, among their other propositions, suggests using a permutation that can be described as $original\ and\ inverse$ order: $ \mathcal{R}_{oi} = \{ \{o_{1}\dots o_{|L|} \} , \{o_{|L|},o_{|L|-1}\dots o_{1}\}\}$.
Following \citealp{zheng2023mcq}, we also
utilize $cyclic$ permutations in which all options are moved in a circular manner between permutations. 
$\mathcal{R}_{cyclic} = \{ \{o_1\dots o_{|L|} \}, \{o_2\dots o_{|L|},o_1\}, \dots ,
\{o_{|L|},o_1\dots \\ o_{|L|-1}\} \}$, where $|\mathcal{R}_{cyclic}|=|L|$.

Finally, we assess the importance of picking these particular option orders by creating random subsets\footnote{Out of the set of possible permutations select random, using \texttt{random.sample} with \texttt{seed = 0}.} $\mathcal{R}_{random2}$ (size $= 2$) and $\mathcal{R}_{randomL}$ (size $=|L|$).

\subsection{Existing metrics}
The central notion of this work is fluctuation, for the measurement of which we adjust the fluctuation rates  metric introduced by \citealp{wei2024unveiling}:

\begin{equation} \label{formula:fr}
FR =  1 - \prod_{\mathclap{\substack{j=1}}}^{|\mathcal{R}|} {\mathbbm{1}{[m_{1}=m_{j}]}}
\end{equation}
By this definition,  we consider a model's answer to fluctuate if at least one response changes throughout permutations. This rigid interpretation allows us to have higher confidence in models' responses.

In the permutation context, one can adapt accuracy by averaging the accuracies obtained in the tested permutations. This change transforms the discrete accuracy into a continuous metric $average \ accuracy$ (which is equivalent to accuracy when computed on $\mathcal{R}_{original}$):
\begin{equation}
AAcc = \frac{1}{|\mathcal{R}|}\sum_{\mathclap{\substack{j=1}}}^{|\mathcal{R}|}{\mathbbm{1}{[m_{j}= a]}}
\end{equation}

Furthermore, we compare the average accuracy results to \textit{strong accuracy}, as introduced by  \citealp{gupta2024changing}, strengthening the accuracy with pairwise comparison of answers across permutations. We update the formula to fit our notation:
\begin{equation} \label{formula:sacc}
    SAcc = \frac{\mathbbm{1}{[m_{1}=a]}}{|\mathcal{R}|} \sum_{\mathclap{\substack{j=1}}}^{|\mathcal{R}|}{\mathbbm{1}[m_{1}=m_{j}]}
\end{equation}

Moreover, we utilize $PriDe$ \cite{zheng2023mcq} in its original implementation by the authors. The method involves computing accuracy using debiased probabilities instead of the original ones. See details about the implementation in the original paper.

To adapt the probability mass of the correct answer to the permutation context, we simply average probabilities across permutations:
\begin{equation}
    Prob = \displaystyle \frac{ 1}{   |\mathcal{R}|} \displaystyle  \sum_{\mathclap{\substack{j=1}}}^{|\mathcal{R}|}  p(l_{a_{j}}|r_{j}).
\end{equation}

We adjust \textit{Brier score} equivalently\footnote{In this work, we convert the metric to \textit{1 - Brier}, to map all the metrics onto the same interval $[0, 1]$ where $0$ is the worst performance and $1$ is the best.}: 
\begin{equation}
     BS = \displaystyle \frac{\displaystyle 1}{\displaystyle  |\mathcal{R}|} \sum_{\mathclap{\substack{j=1}}}^{|\mathcal{R}|} \sum_{l \in L}  \scriptstyle(\displaystyle{\mathbbm{1}}{\displaystyle [l = l_{a_{j}}]}-\  \displaystyle p(l|,r_{j}))^2
\end{equation}

Lastly, we modify the \textit{normalized ENtropy} formula from \citealp{tjuatja2024llms} to incorporate the permutations\footnote{Similarly to $Brier$, we use $1 - Entropy$.}:

\begin{equation}
\begin{multlined}[t]
EN =  \frac{\displaystyle -1}{ |\mathcal{R}| } 
\sum_{\mathclap{\substack{j=1}}}^{|\mathcal{R}|}\sum_{l \in L}\displaystyle  
 \frac{p(l|r_{j}))\cdot log_{2} (p(l|r_{j}))}{log_2(|L|)}
\end{multlined}
\end{equation}

\subsection{Metric proposition} \label{metric_proposition}
Since metrics are averaged across all questions, both average and strong accuracies become hard to interpret. A result of 0.5 can signify both that a model is robust and produces correct answers in all permutations for 50\% of the questions, or that the model is not robust and for all questions there is only a 50\%  chance to get a correct response. We argue that this distinction is important in the context of model reliability, and hence we propose a novel metric, \textit{worst accuracy}, which equals~$1$ iff a model answers correctly throughout all tested permutations:
\begin{equation}
    WAcc = \mathbbm{1}{[m_1}=a] \prod_{\mathclap{\substack{j=1}}}^{|\mathcal{R}|}\mathbbm{1}{[m_{1}=m_{j}]}
\end{equation}
 
One can notice stark similarities between the proposition and Eq.~\ref{formula:sacc}. In fact, the metrics are equal if $|\mathcal{R}|=2$. However, extending the pairwise comparison to include all answers guarantees model robustness on a given question.

\begin{figure*}[h!]
    \centering
    \includegraphics[width=0.9\linewidth]{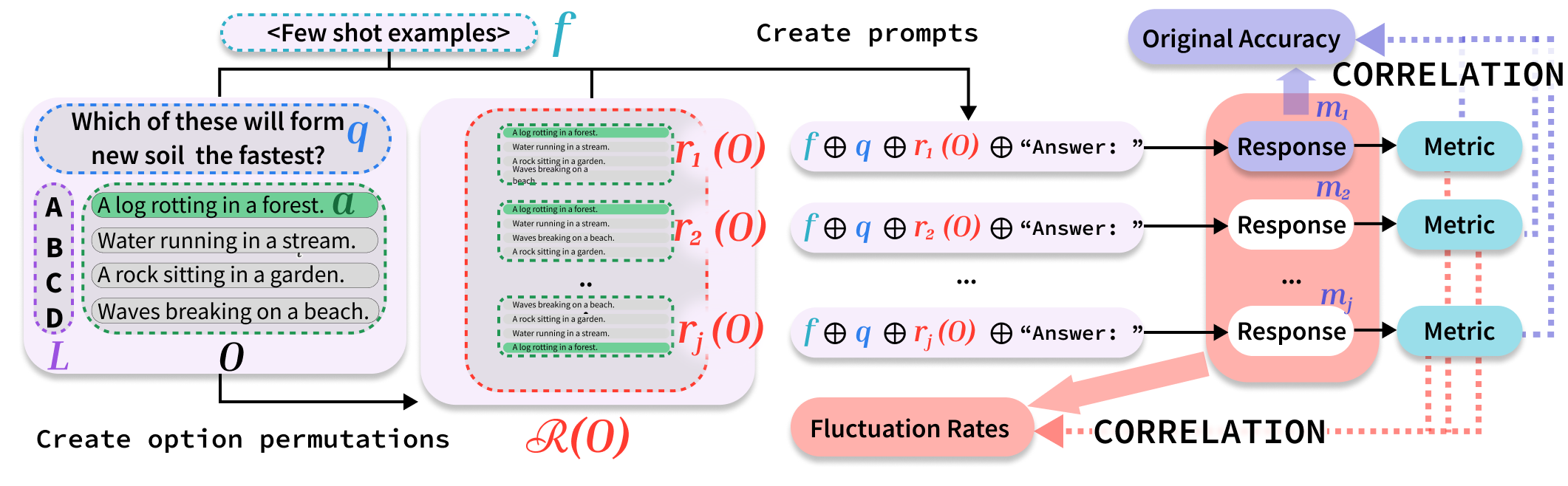}
    \caption{Schematization of the proposed evaluation protocol.}
    \label{explanation}
\end{figure*}
In the original paper \cite{pezeshkpour2023large}, \textit{sensitivity gap}  only receives a textual definition: "difference between the maximum and minimum LLMs’ performance". We provide an interpretation of the metric\footnote{Similarly to $Brier$ and $Entropy$, we use \textit{1 - SensG}.}, using the above-mentioned worst accuracy and an auxiliary metric \textit{best accuracy (BAcc)}, described below.:
 \begin{equation}
     SensG =  BAcc - WAcc
 \end{equation}
\textit{BAcc} considers a question answered if there is at least one permutation in which the model arrives at the correct answer:
\begin{equation}
    BAcc=1 - \prod_{\mathclap{\substack{j=1}}}^{|\mathcal{R}|}{\mathbbm{1}[m_{j}\neq a]}
\end{equation}

\section{Assessment Protocol} \label{protocol}
Having presented all the metrics, one can choose a multitude of assessment protocols.
Since computing all permutations and finding the full fluctuation rates is a costly venture, we argue that an appropriate metric for MCQ evaluation would be highly representative of the full fluctuation rates computed in a lower-cost environment. Therefore, we propose evaluating the correlation of the proposed methodologies with full fluctuation rates. However, a metric should still be illustrative of the model's accuracy on the original option order, since this represents the result of a model on a version it was exposed to. Thus, we additionally propose the following protocol, illustrated in Figure~\ref{explanation}:
\begin{enumerate}
    \item Calculate the accuracy models achieve on the original benchmarks (using the original option order).
    \item Calculate fluctuation rates on all possible permutations of option order for each model and benchmark.
    \item Calculate the metrics from Section \ref{metric_survey} on a smaller subset of permutations for each model on each benchmark.
    \item Find the correlation between metrics and full fluctuation rates using $R^2$.
    \item Find the correlation between metrics and original accuracy using $R^2$.
    \item Find the correlation between a metric and both full fluctuation rates and original accuracy using $R^2$.
\end{enumerate}

\subsection{Models} We perform our experiments on 10 LLMs with parameter sizes below 10B. Models of this size are frequently used for fine-tuning\footnote{At the time of writing 100-900+ fine-tuned versions are available on HuggingFace for each selected model.}, thus making their evaluation more impactful. This size also allows us to perform a costly operation of computing all possible permutations.
In our experiments we use pre-trained and instruct-tuned versions of Llama-3.1-8B \cite{grattafiori2024llama}, Gemma-2-9B \cite{gemmateam2024gemma2improvingopen}, Mistral-7B-v0.3 \cite{jiang2023mistral7b}, Qwen2.5-7B \cite{qwen2025qwen25technicalreport}, as well as R1-Distill-Llama-8B and R1-Distill-Qwen-7B from DeepSeek \cite{deepseekai2025deepseekr1incentivizingreasoningcapability}.  All models are initialized using HuggingFace's \texttt{transformers} library with \texttt{bfloat16} precision.

\subsection{Benchmarks}
Due to potential variability in results coming from slight variations of input text, we choose to use publicly shared Meta's evaluation datasets\footnote{\url{https://huggingface.co/datasets/meta-llama/Llama-3.1-8B-evals}} that contain full final prompts, including instructions, few-shot examples, their order, and option typography for ARC-C \cite{clark2018think}, CSQA \cite{talmor2018commonsenseqa}, MMLU\footnote{The benchmark contains 57 diverse subtasks, in this work we present results from a sample of 12 subtasks.} \cite{hendrycks2020measuring}, AGIEval\footnote{Though originally a 5-option benchmark, AGIEval contains questions with \texttt{nan} as the final option. We remove it and consider such questions to be 4-option, thus creating two subsets AGIEval-4 and AGIEval-5.} \cite{zhong2023agieval}, and Winogrande \cite{sakaguchi2021winogrande}\footnote{See Appendix \ref{benchmark_det} for more information.}.
All benchmarks' prompts can be generalized to the following format: \texttt{"<instruction> <few-shot examples> <test question} $q$ \texttt{> <test options} $r_{j}$\texttt{> Answer: "}. 

\begin{figure*}[h!]

\begin{subfigure}{\textwidth}
    \centering
    \includegraphics[width=0.75\linewidth]{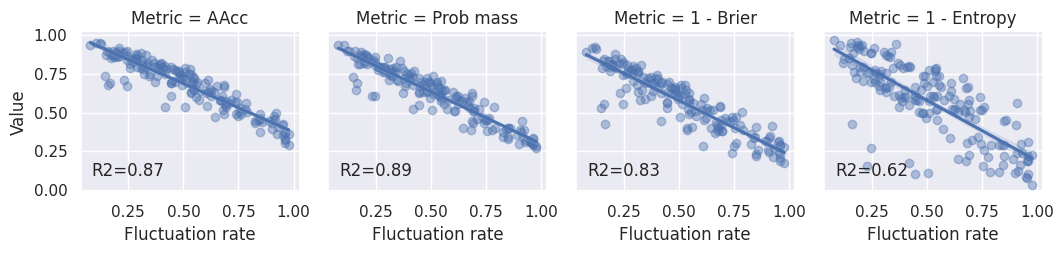}
    \caption{Metrics computed using $\mathcal{R}_{original}$}
    \label{fig:og_metrics}
\end{subfigure}
\\
\begin{subfigure}{\textwidth}
    \centering
    \includegraphics[width=0.75\linewidth]{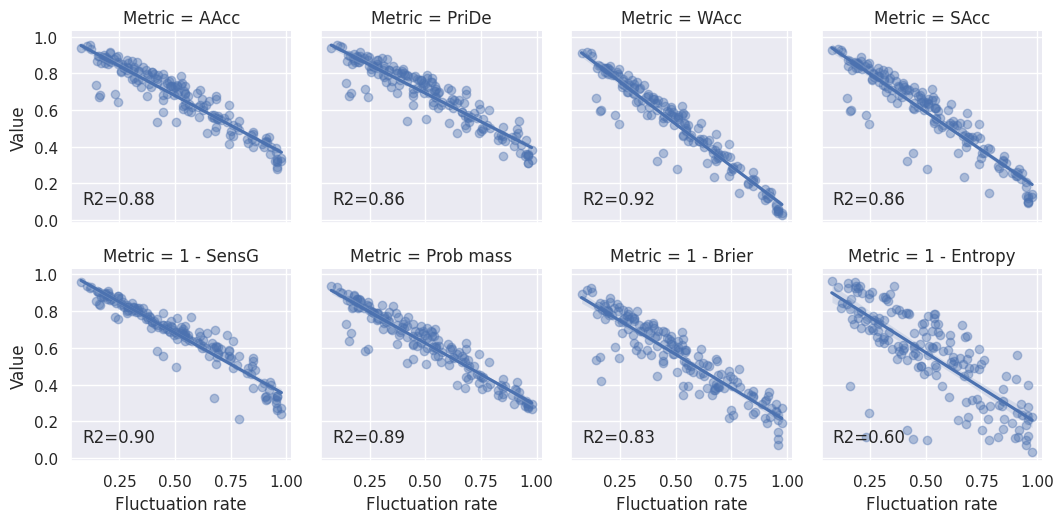}
    \caption{Metrics computed using $\mathcal{R}_{cyclic}$}
    \label{fig:cyclic_metrics}
\end{subfigure}
    \caption{Metrics and full fluctuation rates correlation. Each data point represents results obtained by a model on a benchmark using the given metric.}
    \label{fig:metrics}
\end{figure*}

\section{Results} \label{results}
This section presents the results of Steps 4-6 of the protocol introduced above.
To begin with, we compute the correlation of the metrics with full fluctuation rates using original order and permutation subsets.
 Second, we compare the results when adding correlation with the original accuracy. Lastly, we assess the impact of picking random permutations for metric calculation\footnote{All metrics are computed on the same randomly picked permutations $\mathcal{R}_{random2}$ and $\mathcal{R}_{randomL}$.}. 
\subsection{Correlation with full fluctuation rates}
Figure \ref{fig:og_metrics} shows that all metrics that could be calculated using only the original option order are representative of full fluctuation rates to a great extent, with the probability mass being the best proxy out of the tested metrics. While entropy appears to have the weakest correlation, the $R^2$ measure still indicates a certain level of association. 

Figure \ref{fig:cyclic_metrics} presents the metrics results calculated using each benchmark's cyclic permutations. Interestingly, there is no change in $R^2$ for probability mass and Brier score when adding extra permutations, thus indicating that additional permutations do not contain more information about fluctuation for these metrics. Worst accuracy appears to have the highest correlation with full fluctuation rates on $\mathcal{R}_{cyclic}$. As seen in the plots of the sensitivity gap and strong and worst accuracies, specific data points appear pretty far from the general fit. These points represent the results of models on Winogrande\footnote{Find more detailed representation in Appendix \ref{metrics_appendix}.}, a benchmark with only two options. One potential explanation for this behavior is that the performance of these metrics is dependent on the size of $|L|$ and, therefore, the number of available permutations.

\begin{table*}[h!]
\begin{subtable}{\linewidth}
\centering
\scriptsize
{    \begin{tabular}{llllllllll}
\toprule
 & AAcc & PriDe & WAcc & SAcc & 1 - SensG & Prob mass & 1 - Brier & 1 - Entropy & 1 - FR (partial) \\
 \midrule
$ \mathcal{R}_{oi}$ & 0.873 & 0.863 & 0.870 & 0.870 & 0.640 & \textbf{0.893} & 0.833 & 0.605 & 0.829 \\
$\mathcal{R}_{random2}$ & 0.881 & 0.877 & 0.831 & 0.831 & 0.235 & \textbf{0.894} & 0.836 & 0.594 & 0.479 \\
$\mathcal{R}_{cyclic}$ & 0.877 & 0.863 & 0.923 & 0.863 & 0.896 & 0.894 & 0.832 & 0.602 & \textbf{0.953} \\
$\mathcal{R}_{randomL}$ & 0.880 & 0.868 & 0.914 & 0.864 & 0.866 & 0.894 & 0.835 & 0.600 & \textbf{0.941} \\
\bottomrule
\end{tabular}}
\caption{Target feature = full fluctuation rates.}\label{tab:r2_fluctuation}

\end{subtable}
\\
\\
\begin{subtable}{\linewidth}
\centering
\scriptsize
{    \begin{tabular}{llllllllll}
\toprule
 & AAcc & PriDe & WAcc & SAcc & 1 - SensG & Prob mass & 1 - Brier & 1 - Entropy & 1 - FR (partial) \\
 \midrule
$ \mathcal{R}_{oi}$ & 0.990 &\textbf{ 0.993 }& 0.960 & 0.960 & 0.647 & 0.960 & 0.943 & 0.686 & 0.844 \\
$\mathcal{R}_{random2}$ &\textbf{ 0.979 }& 0.978 & 0.930 & 0.930 & 0.275 & 0.957 & 0.937 & 0.674 & 0.508 \\
$\mathcal{R}_{cyclic}$ & 0.987 & \textbf{0.994} & 0.961 & 0.963 & 0.827 & 0.960 & 0.941 & 0.682 & 0.897 \\
$\mathcal{R}_{randomL}$ & \textbf{0.988} & 0.985 & 0.964 & 0.958 & 0.813 & 0.959 & 0.941 & 0.681 & 0.903 \\
\bottomrule
\end{tabular}}
\caption{Target feature = accuracy on original order.}\label{tab:r2_accs}
\end{subtable}
\\
\\
\begin{subtable}{\linewidth}
\scriptsize
\centering
{\begin{tabular}{llllllllll}
\toprule
 & AAcc & PriDe & WAcc & SAcc & 1 - SensG & Prob mass & 1 - Brier & 1 - Entropy & 1 - FR (partial) \\
\midrule
$ \mathcal{R}_{oi}$ & \textbf{0.932} & 0.928 & 0.915 & 0.915 & 0.643 & 0.927 & 0.888 & 0.645 & 0.836 \\
$\mathcal{R}_{random2}$ & \textbf{0.930} & 0.928 & 0.881 & 0.881 & 0.255 & 0.926 & 0.886 & 0.634 & 0.494 \\
$\mathcal{R}_{cyclic}$ & 0.932 & 0.928 & \textbf{0.942 }& 0.913 & 0.861 & 0.927 & 0.887 & 0.642 & 0.925 \\
$\mathcal{R}_{randomL}$ & 0.934 & 0.926 & \textbf{0.939} & 0.911 & 0.839 & 0.927 & 0.888 & 0.641 & 0.922 \\
\bottomrule
\end{tabular}}
\caption{Target features = full fluctuation rates and original accuracy.}\label{tab:r2_both}

\end{subtable}

\caption{$R^2$ scores for metrics computed on permutation subsets and full fluctuation scores and/or original accuracy. For random subsets, we used the same permutations for all calculations. Best results for each permutation subset are bolded.}\label{tab:1}
\end{table*}

Seeing these results, we investigate if partial fluctuation rates (computed over subsets of permutations) are associated with full fluctuation rates. In fact, such an approach shows the best performance in $\mathcal{R}_{cyclic}$ and $\mathcal{R}_{randomL}$ setups, exceeding the results of the worst accuracy (see Table \ref{tab:r2_fluctuation}). However, such a method appears to be much less stable over just two permutations, with correlation dropping significantly over $\mathcal{R}_{random2}$. Similarly, sensitivity gap performs very poorly on $\mathcal{R}_{random2}$. This can serve as an additional indicator that two permutations are insufficient for calculating these metrics. 

\subsection{Correlation with original accuracy and full fluctuation rates}
As the next step, we find the correlation between the metrics and the accuracy computed on the original benchmark (see the results in Table~\ref{tab:r2_accs}). Though partial fluctuation rates have a substantial correlation with full fluctuation rates, it appears that this strong link comes with less information about original accuracy than other metrics. Similar to the previous results, sensitivity gap and fluctuation rates computed over $\mathcal{R}_{random2}$ demonstrate a drastic drop in comparison to  $ \mathcal{R}_{oi}$, further suggesting the impact of chosen dimensions on the calculation of the metric.

Curiously, the highest correlation with the original accuracy on $ \mathcal{R}_{oi}$ and $\mathcal{R}_{cyclic}$ is achieved by PriDe and not by averaged accuracy. Probability mass, Brier score, worst and strong accuracies are strongly associated with original accuracies, though slightly worse than PriDe and averaged accuracy.

As our final evaluation, we compute the $R^2$ score for correlation with both targets simultaneously (Table \ref{tab:r2_both}). Worst accuracy arises to be the best approach given $\mathcal{R}_{cyclic}$ or $\mathcal{R}_{randomL}$. In contrast, averaged accuracy appears to be the best on $ \mathcal{R}_{oi}$ and $\mathcal{R}_{random2}$, demonstrating the most balanced performance across two target features.

\subsection{Permutation choice impact}
\begin{figure*}[h!]
     \centering
    \includegraphics[width=\linewidth]{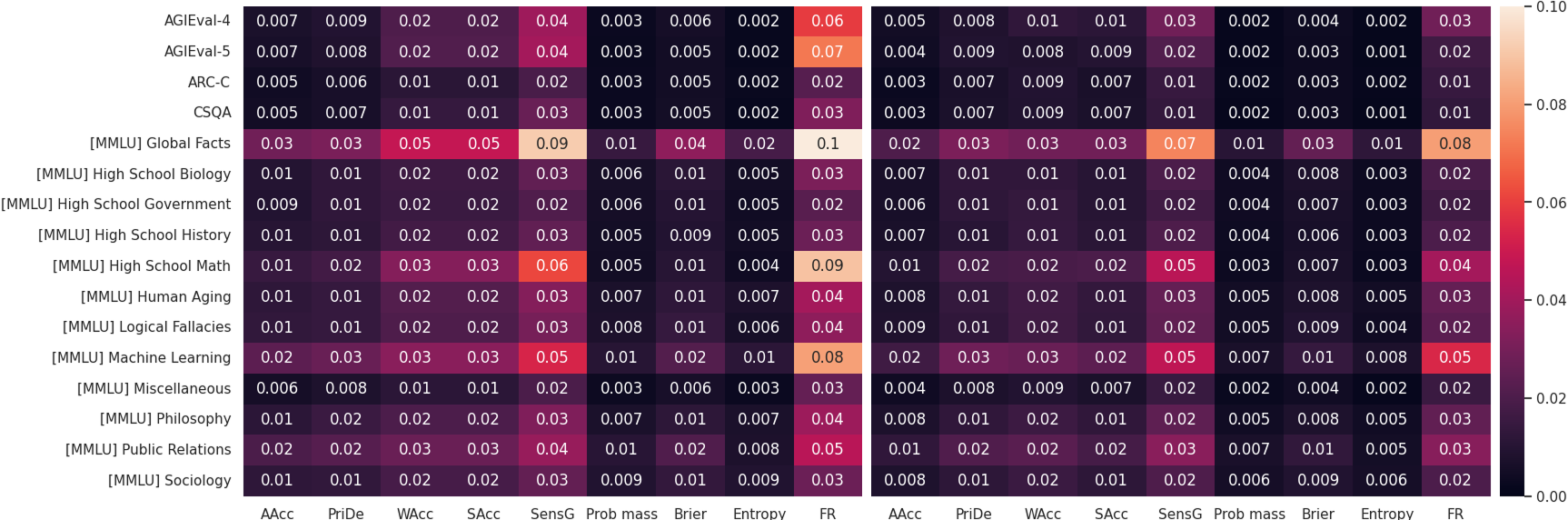}
    \caption{Standard deviation of each metric on a given benchmark, averaged by model. \textit{Left}: standard deviation given random pairs of permutations. \textit{Right:} standard deviation computed on random tuples of permutations of length $|L|$.}
    \label{fig:stds}
 \end{figure*}

Considering the differences in performance when adopting $ \mathcal{R}_{oi}$ and $\mathcal{R}_{random2}$, we compare the standard deviations of the tested metrics. For this purpose, we choose 100 random pairs of permutations for each benchmark except Winogrande\footnote{Since only 2 permutations are available for it.}, as well as 100 random tuples of size $|L|$, and calculate metrics for each of them. We report an averaged standard deviation of a metric on a benchmark in Figure~\ref{fig:stds}.
We find that the standard deviation of the sensitivity gap and partial fluctuation rates computed over random pairs of permutations are the most significant among the metrics, mirroring the observed drops of $R^2$ when replacing $ \mathcal{R}_{oi}$ with $\mathcal{R}_{random2}$. Furthermore, we remark that standard deviations are higher on benchmarks where all models perform worse on the original order\footnote{See the details about models' original accuracies in Appendix \ref{original_accs}.} (e.g. Global Facts, Machine learning, and High School Math). 

Additionally, we notice that within permutations, continuous metrics can increase on some questions, however, to a similar extent decrease on others, and the overall averaged performance stays stable no matter the permutations chosen (reflected by low standard deviation in Figure~\ref{fig:stds}). While this stability allows one to pick random permutations for calculation of the metrics, it appears to be also associated with a capped correlation with fluctuation: $R^2$ values do not improve when adding more permutations (compare Figures \ref{fig:og_metrics} and \ref{fig:cyclic_metrics}). Thus, \textit{computing continuous metrics over several permutations might have no benefit over computing them over $\mathcal{R}_{original}$.}

While using $|L|$ permutations is associated with lower standard deviation, it remains quite significant for PriDe, worst and strong accuracies, sensitivity gap and fluctuation rates. Consequently, selecting random permutations (as proposed in \citealp{gupta2024changing}) might lead to unstable evaluation.

\section{Limitations \& Future Work}

\paragraph{Selection of permutations} As demonstrated in the results, multiple metrics appear sensitive to the permutations chosen to compute them. While we observe this phenomenon, further study is required on the optimal approaches to permutation selection.

\paragraph{Other permutation types} While we illustrated how strongly metrics correlate with fluctuation, we only considered option order permutations. As discussed in Section \ref{related_work}, fluctuation can occur with question paraphrasing, changing option typography, replacing option labels, etc. Further work needs to include these types of permutations in the assessment.

\paragraph{Model sizes} All experiments were performed using similar-sized models. Including models of other sizes is essential to understanding whether the demonstrated correlation of tested metrics is characteristic only of the models of this size or whether a more general pattern exists.

\paragraph{Text generation vs next token prediction} In our experiments, models' answers were decided by the next token with the highest probabilities, but as previous research has demonstrated \cite{wang2024look, wang2024my}, it might be associated with higher fluctuation rates of responses than text generation.  Further research needs to incorporate and analyze both approaches.

\section{Conclusion}
In this paper, we presented a new protocol for metric comparison in the context of answer fluctuation that LLMs exhibit when options of MCQ tasks are permuted. To achieve this, we reviewed, formalized, and computed existing metrics applicable to such benchmarks, and introduced a new metric, worst accuracy.
When applying the evaluation framework, we discovered that:
\begin{enumerate}
    \item Most existing metrics appear to correlate strongly with fluctuation rates.
    \item When only having access to the results of a model on the original order of options, one might employ probability mass for a substantial correlation with full fluctuation rates. However, computing the same metric over multiple permutations does not appear to yield better results.
    \item If information about the original model performance is not of high importance, computing fluctuation rates on cyclic permutations comes to be the best indicator of fluctuation on all possible permutations.
    \item However, if it is essential for the evaluation to represent the original accuracy, the worst accuracy shows the best performance.
\end{enumerate}

Further research is required to extend these findings to different approaches to answer generation by models, a variety of sizes, and other types of permutations that lead to answer fluctuation.

\bibliography{latex/acl_latex_new}
\newpage
\onecolumn
\appendix

\section{Metric Results}

In this section we present detailed results, indicating individual model performance on tested benchmarks. Section \ref{original_accs} demonstrates original accuracies for benchmark pairs.  Section \ref{full_fr} includes full fluctuation rates for model-benchmark pairs. Section \ref{metrics_appendix} presents correlation plots of a metric and full fluctuation rates, detailed by model and benchmark.

\subsection{Original Accuracy} \label{original_accs}
\begin{figure}[h!]
    \centering
    \includegraphics[width=\linewidth]{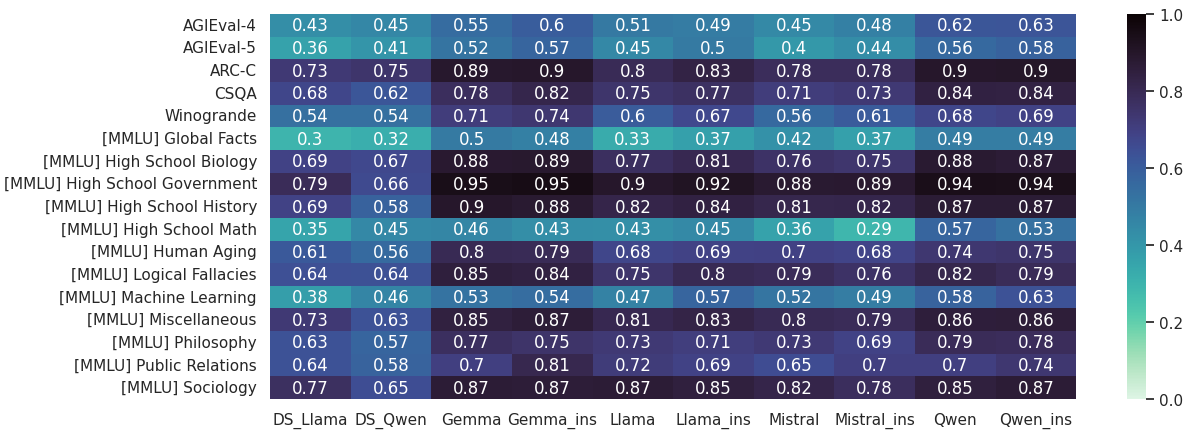}
    \caption{Accuracies obtained by the models on the benchmarks using the original option order.}
    \label{fig:enter-label}
\end{figure}

\subsection{Full Fluctuation Rates} \label{full_fr}
\begin{figure}[h!]
    \centering
    \includegraphics[width=\linewidth]{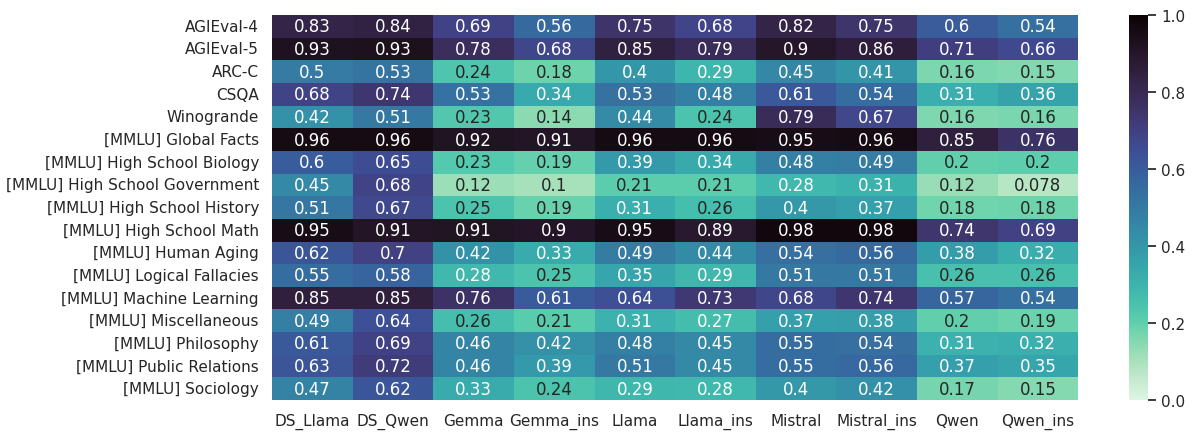}
    \caption{Fluctuation rates of the models on the benchmarks calculated using all permutations.}
    \label{fig:enter-label}
\end{figure}

\newpage
\subsection{Metrics on Different Permutations} \label{metrics_appendix}
\begin{figure}[h!]
    \centering
    \includegraphics[width=0.9\linewidth]{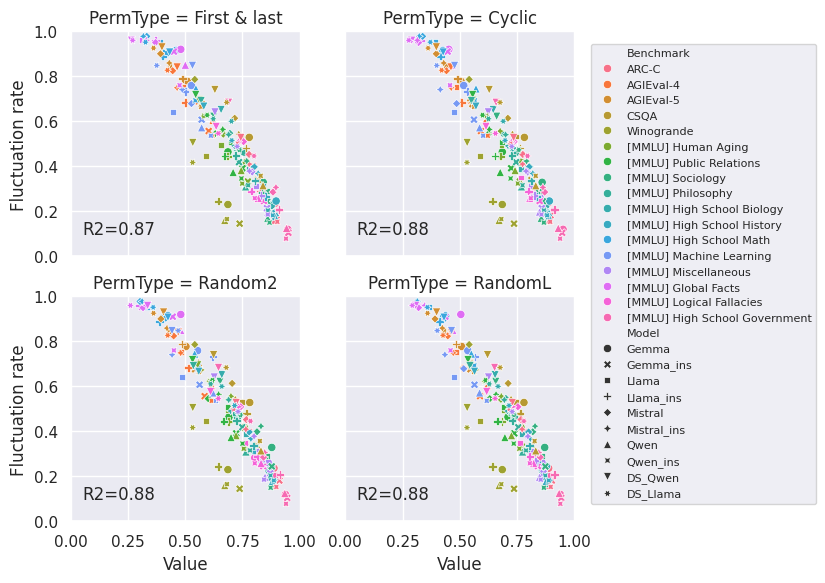}
    \caption{Average accuracy on permutation subsets and full fluctuation rates for all tested models and benchmarks.}
    \label{fig:enter-label}
\end{figure}

\begin{figure}[h!]
    \centering
    \includegraphics[width=0.9\linewidth]{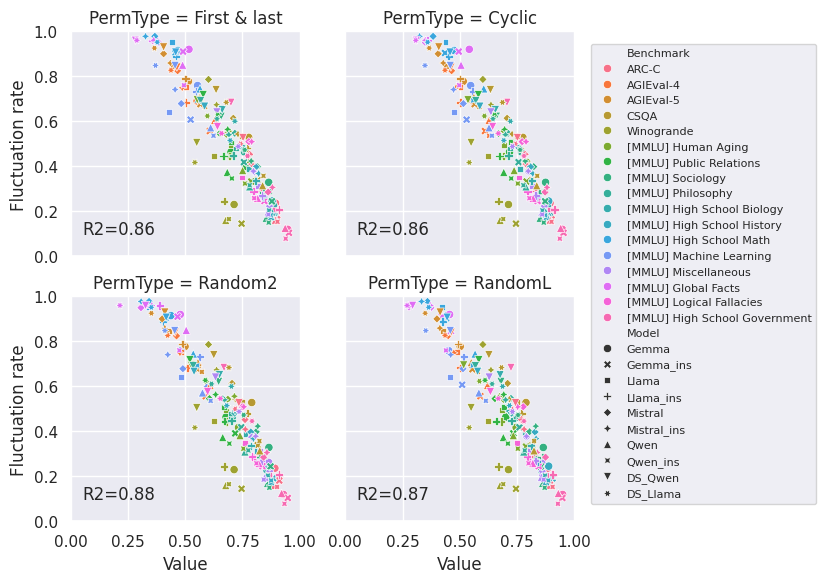}
    \caption{PriDe on permutation subsets and full fluctuation rates for all tested models and benchmarks.}
    \label{fig:enter-label}
\end{figure}
\newpage

\begin{figure}[h!]
    \centering
    \includegraphics[width=0.9\linewidth]{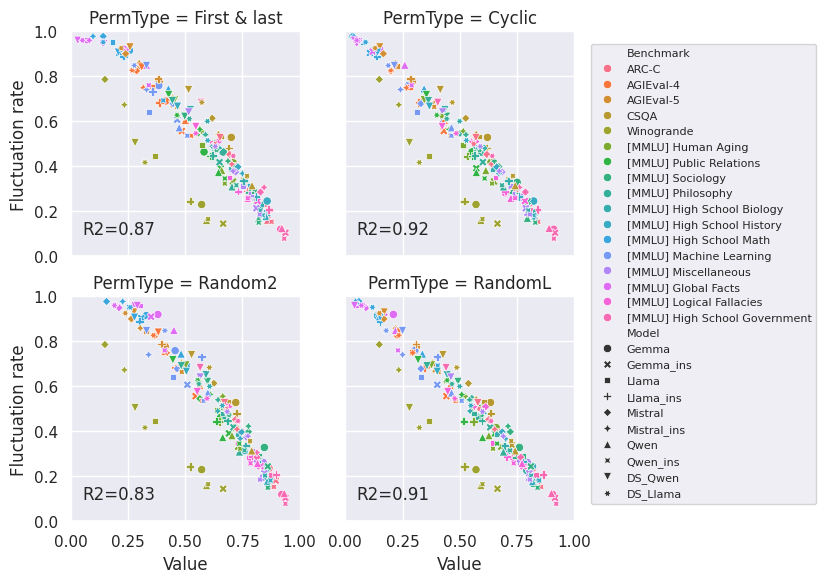}
    \caption{Worst accuracy on permutation subsets and full fluctuation rates for all tested models and benchmarks.}
    \label{fig:enter-label}
\end{figure}

\begin{figure}[h!]
    \centering
    \includegraphics[width=0.9\linewidth]{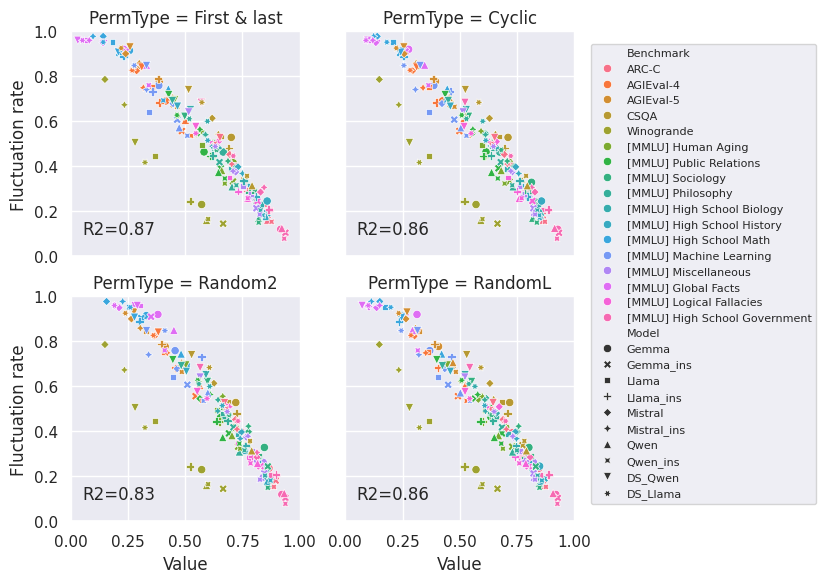}
    \caption{Strong accuracy on permutation subsets and full fluctuation rates for all tested models and benchmarks.}
    \label{fig:enter-label}
\end{figure}
\newpage
\begin{figure}[h!]
    \centering
    \includegraphics[width=0.9\linewidth]{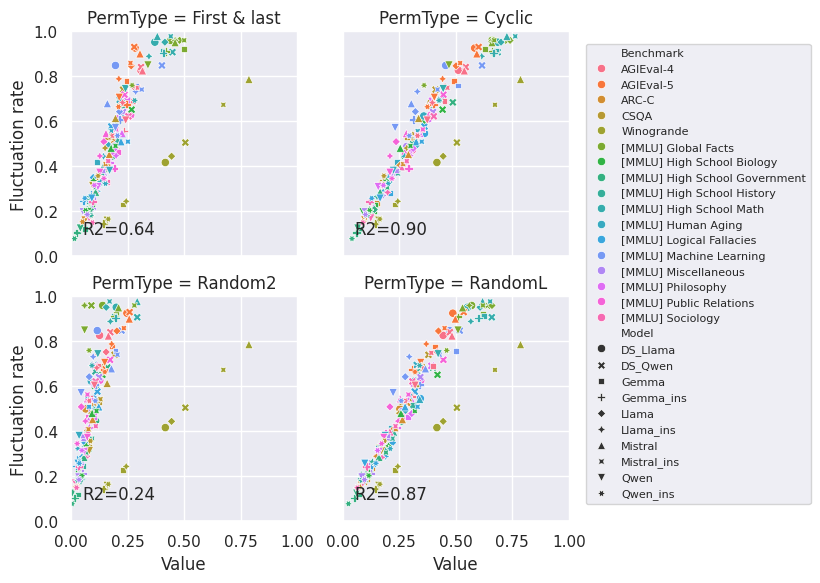}
    \caption{Sensitivity gap on permutation subsets and full fluctuation rates for all tested models and benchmarks.}
    \label{fig:enter-label}
\end{figure}

\begin{figure}[h!]
    \centering
    \includegraphics[width=0.9\linewidth]{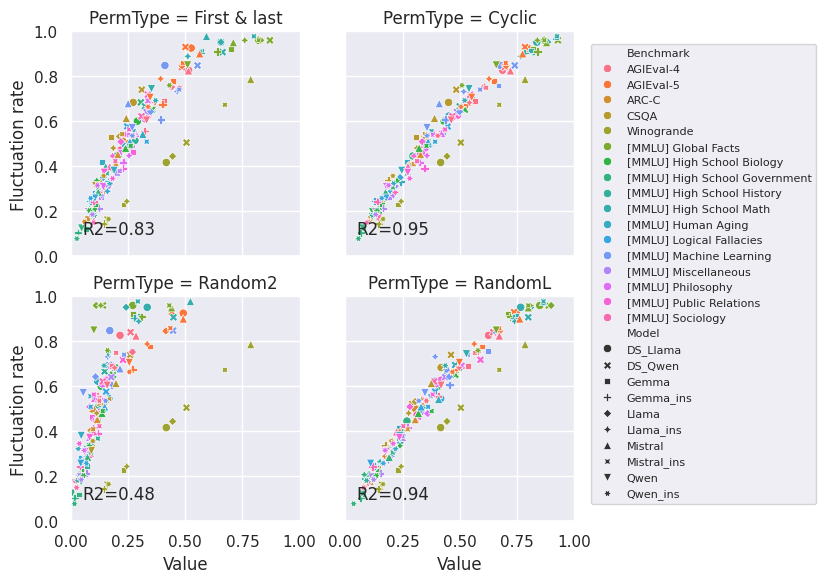}
    \caption{Fluctuation rates on permutation subsets and full fluctuation rates for all tested models and benchmarks.}
    \label{fig:enter-label}
\end{figure}
\newpage
\begin{figure}[h!]
    \centering
    \includegraphics[width=0.9\linewidth]{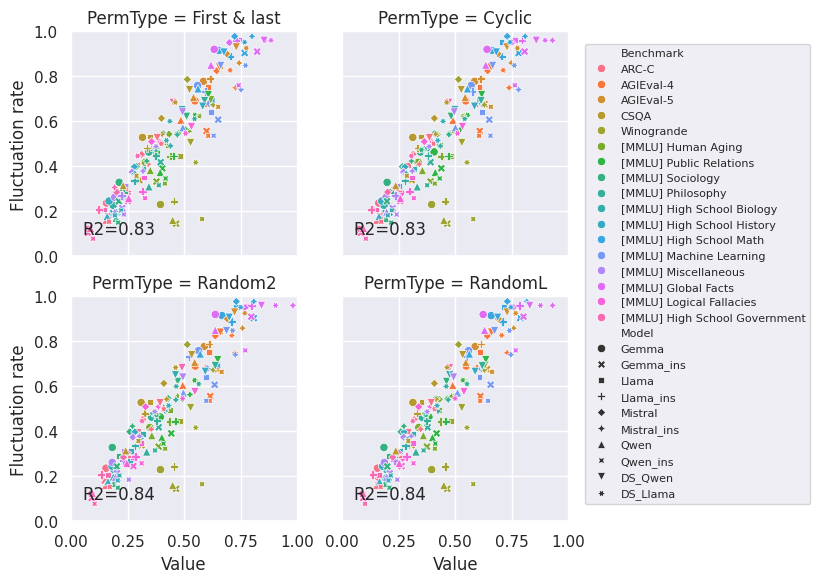}
    \caption{Brier score on permutation subsets and full fluctuation rates for all tested models and benchmarks.}
    \label{fig:enter-label}
\end{figure}

\begin{figure}[h!]
    \centering
    \includegraphics[width=0.9\linewidth]{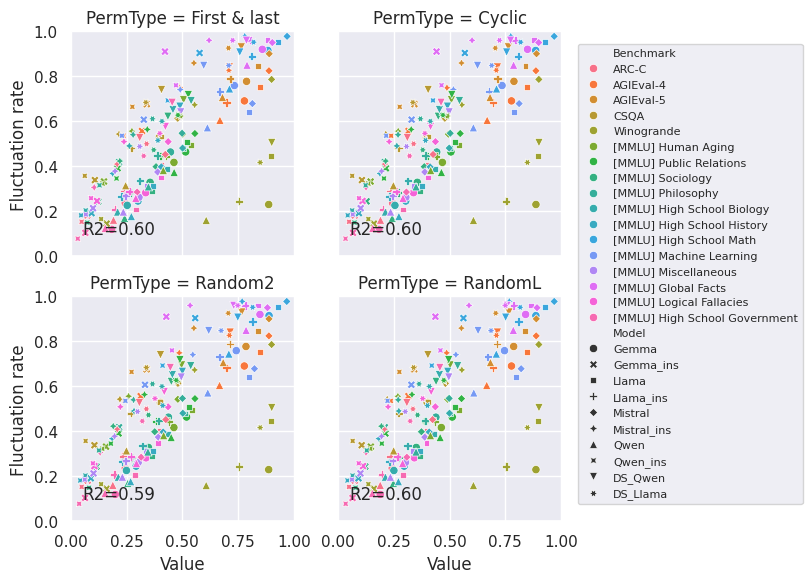}
    \caption{Entropy on permutation subsets and full fluctuation rates for all tested models and benchmarks.}
    \label{fig:enter-label}
\end{figure}

\newpage
\begin{figure}[h!]
    \centering
    \includegraphics[width=\linewidth]{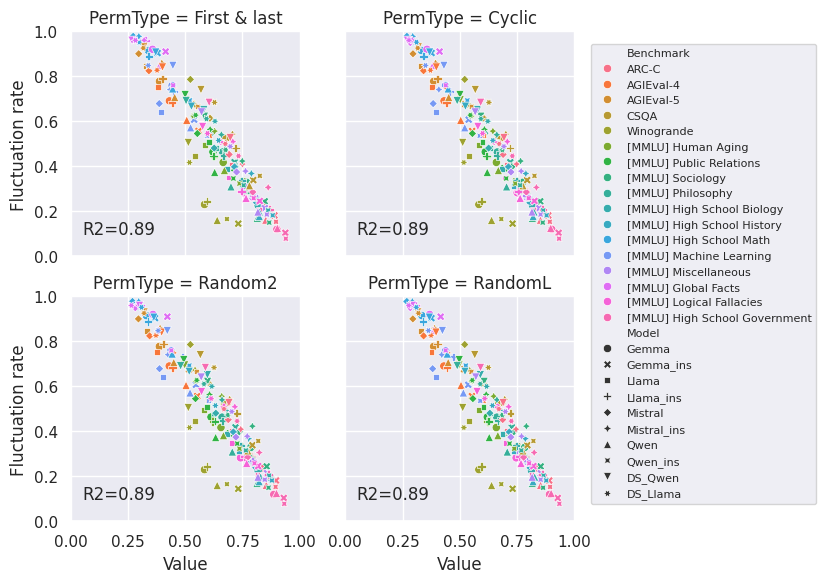}
    \caption{Probability of correct answer on permutation subsets and full fluctuation rates for all tested models and benchmarks.}
    \label{fig:enter-label}
\end{figure}

\newpage
\section{Benchmark Details}
\label{benchmark_det}

\begin{table}[h!]
    \centering
    \small
    \begin{tabular}{cccc}
        \toprule
        \textbf{Benchmark} & \textbf{\# Questions} & \textbf{\# Options}\\
\midrule
ARC-C & 1165 & 4 \\
AGIEval-4 & 1283 & 4 \\
AGIEval-5 & 1263 & 5 \\
CSQA & 1221 & 5 \\
Winogrande & 1267 & 2 \\
MMLU  - Human Aging & 223 & 4 \\
MMLU  - Public Relations & 110 & 4 \\
MMLU  - Sociology & 201 & 4 \\
MMLU  - Philosophy & 311 & 4 \\
MMLU  - High School Biology & 310 & 4 \\
MMLU  - High School History & 204 & 4 \\
MMLU  - High School Math & 270 & 4 \\
MMLU  - Machine Learning & 112 & 4 \\
MMLU  - Miscellaneous & 783 & 4 \\
MMLU  - Global Facts & 100 & 4 \\
MMLU  - Logical Fallacies & 163 & 4 \\
MMLU  - High School Government & 193 & 4 \\
\bottomrule

    \end{tabular}
    \caption{Benchmarks used in the experiments, along with the number of questions in each benchmark and the number of options in each question.}
    \label{tab:benchmarks}
\end{table} 
\end{document}